\documentclass[sigconf,nonacm]{acmart}
\usepackage{cleveref}
\usepackage{siunitx}
\usepackage{algorithm}
\usepackage{algpseudocode}
\Crefname{equation}{Eq.}{Eqs.}
\Crefname{figure}{Fig.}{Figs.}
\Crefname{tabular}{Tab.}{Tabs.}
\Crefname{section}{Sec.}{Secs.}
\Crefname{algorithm}{Alg.}{Algs.}

\AtBeginDocument{%
  \providecommand\BibTeX{{%
    \normalfont B\kern-0.5em{\scshape i\kern-0.25em b}\kern-0.8em\TeX}}}

\setcopyright{acmcopyright}
\copyrightyear{2018}
\acmYear{2018}
\acmDOI{10.1145/1122445.1122456}

\acmDOI{10.1145/nnnnnnn.nnnnnnn} % To be updated after completing copyright process
\acmISBN{978-x-xxxx-xxxx-x/YY/MM} % To be updated after completing copyright process
\acmConference[GECCO '22]{The Genetic and Evolutionary Computation Conference 2022}{July 9--13, 2022}{Boston, USA}
\acmYear{2022}
\copyrightyear{2022}

\begin{document}

\title{Coefficient Mutation in the Gene-pool Optimal Mixing Evolutionary Algorithm for Symbolic Regression}


\author{Marco Virgolin}
\affiliation{%
  \institution{Centrum Wiskunde \& Informatica}
  \city{Amsterdam}
  \country{the Netherlands}}
\email{marco.virgolin@cwi.nl}

\author{Peter A. N. Bosman}
\affiliation{%
  \institution{Centrum Wiskunde \& Informatica}
  \city{Amsterdam}
  \country{the Netherlands}}
\email{peter.bosman@cwi.nl}

\renewcommand{\shortauthors}{Virgolin and Bosman}

\begin{abstract}
  Currently, the genetic programming version of the gene-pool optimal mixing evolutionary algorithm (GP-GOMEA) is among the top-performing algorithms for symbolic regression (SR).
  A key strength of GP-GOMEA is its way of performing variation, which dynamically adapts to the emergence of patterns in the population.
  However, GP-GOMEA lacks a mechanism to optimize coefficients.
  In this paper, we study how fairly simple approaches for optimizing coefficients can be integrated into GP-GOMEA. In particular, we considered two variants of Gaussian coefficient mutation. We performed experiments using different settings on 23 benchmark problems, and used machine learning to estimate what aspects of coefficient mutation matter most.
  We find that the most important aspect is that the number of coefficient mutation attempts needs to be commensurate with the number of mixing operations that GP-GOMEA performs.
  We applied GP-GOMEA with the best-performing coefficient mutation approach to the data sets of SRBench, a large SR benchmark, for which a ground-truth underlying equation is known.
  We find that coefficient mutation can help re-discovering the underlying equation by a substantial amount, but only when no noise is added to the target variable.
  In the presence of noise, GP-GOMEA with coefficient mutation discovers alternative but similarly-accurate equations.
\end{abstract}




\maketitle

\section{Introduction}
Symbolic regression (SR) is the task of discovering a governing mathematical equation that underlies the given data~\cite{koza1994genetic}.
Algorithms implementing SR can be of wildly different nature, including exhaustive or greedy search strategies~\cite{mcconaghy2011ffx,kammerer2020symbolic,de2018greedy,rivero2022dome}, genetic programming (GP) and other evolutionary approaches~\cite{koza1994genetic,schmidt2009distilling,kantor2021simulated}, deep neural networks~\cite{petersen2019deep,biggio2021neural,d2022deep}, as well as hybrids~\cite{cranmer2020discovering} and pipelines~\cite{udrescu2020ai}.

A recent, large benchmark study called SRBench has demonstrated that GP-based algorithms are among the best approaches for tackling SR~\cite{la2021contemporary}.
Generally, GP works by initializing a random population of candidate solutions and improving this population in an iterative fashion, by means of component recombination between parent solutions, mutation, and stochastic survival of the fittest~\cite{koza1994genetic}.
At the time of writing, one of the top-performing algorithms in SRBench is the GP version of the gene-pool optimal mixing evolutionary algorithm (GP-GOMEA)~\cite{virgolin2017scalable,virgolin2021improving}, which strikes a good balance in terms of delivering small but accurate solutions.
In fact, obtaining small, simple solutions is important in SR to enhance the chance that these solutions will be interpretable~\cite{virgolin2022less} (else, one may as well use uninterpretable methods such as deep neural networks for non-linear regression~\cite{kadra2021well}). 

A key strength behind the performance of GP-GOMEA is its particular form of variation, which is not completely at random as is the case, e.g., for classic subtree crossover~\cite{koza1994genetic}.
Every generation, GP-GOMEA infers a statistical model of promising crossover masks based on the emergence of component patterns, and then uses the crossover masks to perform mixing operations.
While GP-GOMEA has been shown capable of performing particularly well, there is no mechanism included that is aimed at optimizing coefficients (i.e., constants) that appear in the solutions.
In fact, coefficients are sampled at random during initialization, and only swapped among solutions during variation.
This means that, for GP-GOMEA to obtain a certain coefficient that has not been sampled at initialization (e.g., $2.3$), a pattern of components need to be opportunely assembled (e.g., $1.2 \times 2.0 - 0.1$).
Hence, reaching a specific coefficient with high numerical precision is unlikely, and even obtaining a coarse approximation might be inefficient.

In this paper, we consider simple evolution-based approaches for coefficient optimization in GP, and evaluate different ways of integrating them in GP-GOMEA.
In particular, we consider two types of Gaussian mutations, one inspired from temperature decay in simulated annealing~\cite{van1987simulated} and one from evolution strategies~\cite{beyer2002evolution}, and assess at what point during GP-GOMEA such mutations should be applied, with what probability, and for how many attempts. 
We opted for random coefficient mutation instead of gradient descent, which is leveraged in many machine learning algorithms~\cite{bottou2010large}, because the former does not require differentiability and is therefore more generally applicable. Moreover, it represents a reasonable starting point for a first study on including coefficient optimization in GP-GOMEA, and can be used as a baseline for comparing more involved approaches in the future (e.g., the Levenberg-Marquardt algorithm~\cite{more1978levenberg}, which is adopted within GP in~\cite{kommenda2020parameter}).
We remark that findings on coefficient optimization in classic GP (such as in~\cite{kommenda2020parameter,dick2020feature}) do not necessarily apply to GP-GOMEA because variation in GP-GOMEA is very different than in classic GP (see \Cref{sec:gp-gomea}).

The remainder of this paper is organized as follows.
In \Cref{sec:background}, we formalize the problem setting of SR, provide a salient background on GP-GOMEA, and report on related work.
In \Cref{sec:coeff-mut}, we introduce the coefficient mutation approaches considered here.
\Cref{sec:experimental-setup} describes the experimental setup.
In \Cref{sec:first-exp} we perform an experiment on 23 benchmark data sets taken from those considered in~\cite{oliveira2018analysing}, aimed at understanding what coefficient mutation approach is most promising in GP-GOMEA.
Then in~\Cref{sec:second-exp}, we apply our findings to the data sets of SRBench for which the data-generating equation is known (so-called \emph{Feynman} and \emph{Strogatz} data sets), to assess whether coefficient mutation can help GP-GOMEA to recover the true underlying equation.
\Cref{sec:discussion} contains a discussion on the overall findings of this paper and \Cref{sec:conclusion} concludes it.

\section{Background}
\label{sec:background}

\subsection{Symbolic regression}
In SR, we are given a data set $\mathcal{D} = \{ \left( \mathbf{x}_i, y_i \right)\}^n_{i=1}$ where $i$ is the index of an observation, $\mathbf{x}_i = \left( x^{(1)}_i, x^{(2)}_i, \dots x^{(d)}_i \right)^\top \in \mathbb{R}^d$ is a vector of $d$ feature values, and $y_i \in \mathbb{R}$ is the target variable (also called dependent variable or label).
We are tasked with finding a function $f$ from a family of functions $\Omega$ such that, $\forall i, f(\mathbf{x}_i) \approx y_i$.

The difference between SR and traditional regression lies in $\Omega$.
In the latter, we choose $\Omega$ to contain functions that can only be different in terms of numerical coefficients $\Theta = (\theta_1, \dots, \theta_k)^\top \in \mathbb{R}^k$.
For example for linear regression, $k=d+1$ and $\Omega$ contains all functions of the form $f_\Theta(\mathbf{x}) := \theta_1 x^{(1)} + \theta_2 x^{(2)} + \dots + \theta_d x^{(d)} + \theta_{d+1}$.

In SR, besides $\Theta$, $\Omega$ is defined in terms of a set of atomic functions, hereafter denoted by $\mathcal{F}$. 
For example, a possible choice of $\mathcal{F}$ is $\{ +, -, \times, \div, \sin, \cos, \exp, \log \}$.
A function in $\Omega$ is then any function that can be obtained by composition of the atomic functions in $\mathcal{F}$, together with the features and arbitrary numerical coefficients.
For example for the choice of $\mathcal{F}$ made above, $f_a \left(x^{(1)}, x^{(2)} \right) := x^{(1)} \times \sin x^{(2)} - 4$ belongs to $\Omega$, while $f_b \left(x^{(1)} \right) := \sqrt{x^{(1)}}$ does not.

\subsection{GP-GOMEA}
\label{sec:gp-gomea}
GP-GOMEA operates as many other GP algorithms, i.e., by iterative improvement of a population of candidate solutions which are typically initially randomly generated. 
However, GP-GOMEA also has some notable differences.
The overall workings of GP-GOMEA are displayed in \Cref{alg:gp-gomea}, while a detailed description is given in the following sections.

\begin{algorithm}
\caption{Overall workings of GP-GOMEA}\label{alg:gp-gomea}
\begin{algorithmic}[1]
\State $\mathcal{P} \gets \text{InitializePopulation}()$ \emph{\# See \Cref{sec:representation}}
\While{$\text{BudgetLeft}()$}
    \State $\text{FOS} \gets \text{BuildFOS}(\mathcal{P})$ \emph{\# See \Cref{sec:fos}}
    \State $\mathcal{O} \gets \emptyset $
    \For{$p_i \in \mathcal{P}$}
        \State $o_i \gets \text{GOM}(p_i, \text{FOS}, \mathcal{P})$ \emph{\# See \Cref{sec:gom}}
        \State $\mathcal{O} \gets \mathcal{O} \cup \{o_i\}$
    \EndFor
    \State $\mathcal{P} \gets \mathcal{O}$
\EndWhile
\end{algorithmic}
\end{algorithm}

\subsubsection{Representation}
\label{sec:representation}
\begin{figure}
    \centering
    \includegraphics[width=0.8\linewidth]{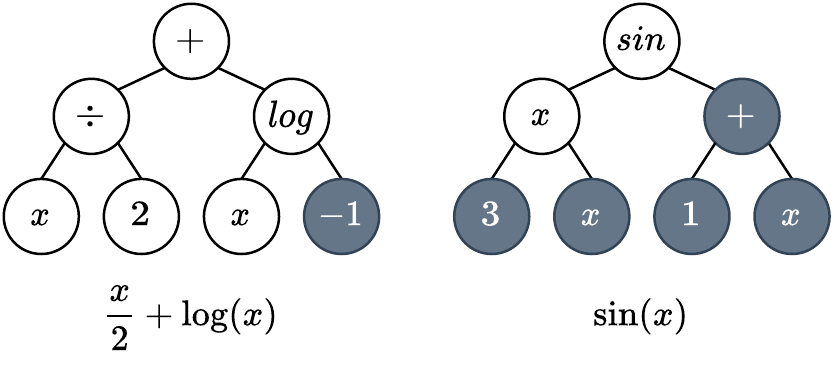}
    \vspace{-.3cm}
    \caption{Two examples of the tree-based encoding of GP-GOMEA for maximal depth $2$ and maximal arity $2$. Gray nodes are introns.}
    \label{fig:gp-gomea-tree}
\end{figure}

Like classic GP~\cite{koza1994genetic}, GP-GOMEA performs SR by encoding the evolving equations with trees.
Internal nodes implement the atomic functions, while leaf nodes implement features or coefficients (constants).
However, in GP-GOMEA all trees adhere to a same \emph{template}; see \Cref{fig:gp-gomea-tree}.
Given a maximal depth, the template consists of a full $m$-ary tree, where $m$ is the maximal arity (i.e., number of input arguments) among the atomic functions in $\mathcal{F}$.
Not all nodes in the template are always \emph{active} , i.e., there can be introns.
Namely, for a node implementing a function of arity $a < m$, only the left-most $a$ child nodes are used as input arguments (features and coefficients are considered to have null arity).

\subsubsection{Family of subsets}
\label{sec:fos}
The reason why GP-GOMEA enforces trees to fit a template is to allow a positional recombination akin to genetic algorithms, which naturally fits the implementation of a statistical model representing promising crossover masks.
Let $\ell$ be the maximum number of nodes in the template tree used in GP-GOMEA. 
Then, in an fixed tree parsing order, any node position is uniquely identified with an index in $\{1, 2, \dots, \ell\}$. 
For example using pre-order parsing, index $1$ identifies the root node, index $2$ the left-most child of the root node, and $\ell$ the right-most leaf.
We can now define the \emph{family of subsets} (FOS), which is a set of sets (called subsets) which, in turn, contain node indices.
For example, a possible FOS is $\{ \{1\}, \{3\}, \{1, 2\}, \{1, 2, 3 \} \}$, with $\{1,2\}$ being one of the subsets of this FOS.
GP-GOMEA uses the FOS to perform mixing operations, using each subset as a crossover mask.
For example the subset $\{3\}$ prescribes to generate an offspring by changing the 3rd node, while  $\{1,2\}$ prescribes to change the first and second node, importantly, \emph{at the same time}.

Now, under the hypothesis that encoding positions that exhibit strong inter-dependencies should be recombined jointly (i.e., as building blocks), the default FOS of GP-GOMEA is the \emph{linkage tree} (LT)~\cite{thierens2010linkage}. 
The LT is built by (1) measuring the mutual information between pairs of node positions (each node position is seen as a random variable, and the population is seen as a sample from this random variable), and (2) approximating higher-order interaction levels by means of hierarchical clustering. 
This results in a tree-like structure, hence the name ``LT'' (not to be confused with the trees used to represent solutions). 
Implementation details, including a normalization step to account for the lack of uniformity of the distribution in the initial population of GP, are reported in~\cite{virgolin2021improving}.

\subsubsection{Gene-pool optimal mixing}
\label{sec:gom}
The gene-pool optimal mixing (GOM) variation operator is used to generate an offspring from a parent solution.
The pseudo-code of GOM is presented in \Cref{alg:gom}; the strategies of coefficient mutation included in the pseudo-code are described later, in \Cref{sec:strategies}.
GOM is applied to every population member. First, a clone of the solution is created. Then, GOM performs as many steps, or mixing attempts, as the number of subsets in the FOS to improve this clone. 
For each subset of the FOS (considered in random order), a GOM step consists of (1) considering the nodes identified by the subset (e.g., $\{1,4,5\}$ identifies the first, fourth, and fifth node, according to the chosen tree parsing order), (2) picking a random member of the population to act as donor, and (3) replacing the nodes identified by the subset in the offspring with the corresponding nodes from the donor.
If this mixing attempt leads to equal or better fitness, the change is kept, else, the change is rolled back (see \Cref{alg:assess-changes}).
For efficiency, the fitness is evaluated only when the mixing attempt is \emph{meaningful}: evaluation is not performed when the nodes copied from the donor represent the same functions as those already present in the offspring, or the nodes copied from the donor end up in intron positions and thus do not influence the computations carried out by the tree (or a combination of the two).

Note that since the LT contains $2 \ell - 2$ subsets (details in~\cite{virgolin2021improving}) and a fitness evaluation is performed after every meaningful recombination attempt, GP-GOMEA typically performs many more fitness evaluations per generation than classic GP (where each solution is evaluated once).
As mentioned in the introduction, this different way of performing recombination calls for an assessment of how coefficient optimization can best be integrated in GP-GOMEA.

\begin{algorithm}
\caption{GOM, including the different strategies we investigated for when to apply coefficient mutation (in {\color{OliveGreen}green})}\label{alg:gom}
\begin{algorithmic}[1]
\Function{GOM}{$p, \text{FOS}, \mathcal{P}$}
\State $o \gets \text{Clone}(p)$ \emph{\# Clones components and also fitness}
\For{$S \in \text{FOS}$} \emph{\# Subsets considered in random order}
    \State \emph{\# Beginning of a GOM step}
    \State $d \gets \text{SampleRandomDonor}(\mathcal{P})$
    \State $o^\prime \gets \text{InheritNodesBySubset}(o, d, S)$
    \color{OliveGreen}
    \State \emph{\# Strategy 4 is applied within a GOM step}
    \If{Strategy 4}
        \State $o^\prime \gets \text{CoefficientMutationToNodesBySubset}(o^\prime, S)$
    \EndIf
    \color{black}
    \State $o \gets \text{AssessChangesAndReturnBest}(o^\prime, o)$ \emph{\# \Cref{alg:assess-changes}}
    \State \emph{\# End of a GOM step}
    \color{OliveGreen}
    \State \emph{\# Strategy 3 is applied in between GOM steps}
    \If{Strategy 3}
        \State $o^\prime \gets \text{CoefficientMutation}(o)$
        \State $o \gets \text{AssessChangesAndReturnBest}(o^\prime, o)$ \emph{\# \Cref{alg:assess-changes}}
    \EndIf
    \color{black}
\EndFor
\color{OliveGreen}
\State \emph{\# Strategies 1 and 2 are applied after GOM has been completed}
\If {Strategy1 or Strategy2}
    \State $k \gets 1 \textbf{ if } \text{Strategy1}, \textbf{ else } |\text{FOS}|$
    \For {$i \in 1,\dots,k$}
        \State $o' \gets \text{CoefficientMutation}(o)$
        \State $o \gets \text{AssessChangesAndReturnBest}(o^\prime, o)$ \emph{\# \Cref{alg:assess-changes}}
    \EndFor
\EndIf
\color{black}
\State \Return $o$
\EndFunction

\end{algorithmic}
\end{algorithm}

\begin{algorithm}
\caption{Auxiliary function used to decide whether changes should be kept or discarded}
\label{alg:assess-changes}
\begin{algorithmic}[1]
\Function{AssessChangesAndReturnBest}{$o^\prime, o$}
\If {$\text{NoMeaningfulChange}(o,o^\prime)$} \emph{\# E.g., introns}
    \State \Return $o^\prime$ \emph{\# Changes led to same fitness}
\EndIf
\State $\text{Fitness}(o^\prime) \gets \text{EvaluateFitness}(o^\prime)$ \emph{\# Evaluation needed}
\If{$\text{Fitness}(o^\prime) \geq \text{Fitness}(o)$} \emph{\# Assume maximization}
    \State \Return $o^\prime$ \emph{\# Changes led to equal or better fitness}
\EndIf
\State \Return $o$ \emph{\# Changes led to worse fitness}
\EndFunction
\end{algorithmic}
\end{algorithm}

\subsection{Related work}
A number of works has considered coefficient optimization in GP.
An early approach is~\cite{howard1995ga}, where a genetic algorithm is integrated within GP, to optimize the coefficients.
Because of its efficiency, many works use differentiable atomic functions and (variants of) gradient-descent.
One of such works is~\cite{topchy2001faster}, where gradient descent is integrated in tree-based GP, and tested in a Baldwinian or Lamarckian fashion.
In~\cite{kommenda2020parameter}, instead of gradient descent, the Levenberg-Marquardt algorithm is adopted.
\cite{zhang2005learning} and \cite{izzo2017differentiable} use gradient descent to optimize multiplicative coefficients placed at the edges connecting nodes, respectively in tree-based and Cartesian genetic programming~\cite{miller2006redundancy}.
A similar formulation is considered in~\cite{trujillo2014evaluating}, using a trust region method to optimize the coefficients.
Lastly,~\cite{la2018learning} uses a differentiable multi-tree representation where gradient descent optimizes inter- and intra-tree coefficients.
Several works have proposed to optimize coefficients using (different types of) coefficient mutation, typically in tree-based GP~\cite{evett1998numeric,babovic2000genetic,langdon2000seeding,hein2018interpretable}.
The hybrid neural-evolutionary approach in~\cite{cranmer2020discovering} includes coefficient optimization with both dedicated mutations and the Nelder-Mead algorithm~\cite{nelder1965simplex}.
In this work we consider two versions of coefficient mutation in GP-GOMEA; to the best of our knowledge, coefficient optimization has not been attempted before in GP-GOMEA.

\section{Coefficient mutation}
\label{sec:coeff-mut}
We use the traditional representation of numerical coefficients by which they are implemented as leaf nodes returning a constant output.
These constant nodes are generated during population initialization, alongside leaf nodes that represent problem variables (in the case of SR, features of the data set).
We consider the use of \emph{ephemeral random constants}, i.e., constant nodes whose value is decided at the moment of their instantiation, according to the chosen distribution~\cite{poli2008field}.
Traditionally, GP-GOMEA does not include subtree or one-point mutation, thus random initialization is solely responsible for what coefficients will be available for the entire evolutionary process (however, a version of GOMEA for grammatical evolution includes mutation~\cite{medvet2018gomge}).

We consider two coefficient mutation approaches that update a current constant value $c$ with the rule:
\begin{equation}
    c^\prime = \delta(c, \mathcal{H}),
\end{equation}
where $\mathcal{H}$ is a set of $k$ hyper-parameters and $\delta : \mathbb{R}^{1+k} \rightarrow \mathbb{R}$ is the update rule.
We remark that having the update rule depend on $c$ means that the new value of the coefficient will depend on the previous value, as opposed to be generated irrespective of it, e.g., as done when replacing a constant leaf node with another one in one-point mutation.

In the next sections, we describe the two ways of implementing $\delta$ and the strategies to integrate coefficient mutation in GP-GOMEA that we considered.

\subsection{Coefficient mutation types (how to optimize)}
\subsubsection{Evolution strategy-like} 
The first approach we consider is inspired by the self-adaptation mechanism in evolution strategies (ES)~\cite{beyer2002evolution}.
Each constant node contains a \emph{meta-parameter} $\sigma$ that is specific to that node, and is initialized with:
\begin{equation}
\label{eq:sigma}
    \sigma = \max \left( \exp(\mathcal{N}(0,\gamma^2)), \epsilon \right),
\end{equation}
with $\mathcal{N}(0,1)$ being the normal distribution with null mean and unit variance, and $\gamma$ and $\epsilon$ being real-valued hyper-parameters common to all constant nodes.
In this paper, we fix $\gamma=0.1$ and $\epsilon=10^{-16}$.

The update rule for the coefficient the constant node represents is then:
\begin{equation}
\label{eq:es-update-c}
    c^\prime = \mathcal{N}(c, \sigma^2), 
\end{equation}
and, importantly, when the update is triggered, then also $\sigma$ is updated, with:
\begin{equation}
\label{eq:es-update-sigma}
    \sigma^\prime = \max\left( \sigma \exp(\mathcal{N}(0,\gamma^2)), \epsilon \right).
\end{equation}
The intuition is that constant nodes will implicitly evolve the update parameter $\sigma$ to become appropriate, e.g., for two nodes with the right coefficient $c$, the node with smaller $\sigma$ will be more likely to lead to a good fitness and survive.
We call this approach ES-like.

\subsubsection{Temperature-based}
The second approach we consider is the one used in~\cite{hein2018interpretable}, i.e.,
\begin{equation}
\label{eq:temp-based}
    c^\prime = \mathcal{N}(c, (c\tau)^2),
\end{equation}
where $\tau \in \mathbb{R}$ is a hyper-parameter, which we call \emph{temperature}, common to all constant nodes.
Note that with this \emph{temperature-based} approach, coefficients with larger magnitude will necessarily have larger mutations than coefficients with smaller magnitude.

While the ES-like approach has an adaptive way of sampling thanks to $\sigma$ being implicitly evolved, the same is not true here.
We therefore experiment with ways to update $\tau$ over the course of the evolution.
In particular, we define the hyper-parameters of \emph{decay} $d$ and \emph{patience} $t$, inspired from simulated annealing~\cite{van1987simulated} and learning rate annealing, which is common in deep learning.
The decay $d \in (0, 1)$ updates $\tau$ by $\tau^\prime = \tau \times d$ when $t$ is reached, where $t$ is the number of consecutive generations in which a better elitist solution has not been found.

\subsection{Coefficient mutation strategies (when and for how long to optimize)}
\label{sec:strategies}
We set up coefficient mutation to work as follows.
When coefficient mutation is \emph{applied} to a solution, we consider all the coefficients (in our case, constant nodes).
For each coefficient, we sample whether it should be mutated, akin to one-point mutation.
The probability of mutating a coefficient is a hyper-parameter; if, e.g., set to 0.5, in expectation half of the coefficients of a solution will be mutated, while the other half will remain to their current value.
Note that as per our wording, \emph{applying} coefficient mutation can result in no changes (e.g., when the probability of mutation is set to be small).

What is left to decide is when and how many times to apply coefficient mutation.
We propose the following strategies (see \Cref{alg:gom}):
\begin{enumerate}
    \item \emph{Once, after GOM}: This strategy resembles the way coefficient optimization is often applied in tree-based GP, i.e., only after subtree crossover, subtree mutation, one-point mutation, reproduction, or other recombination operators have taken place. 
    With this strategy we do something similar: we apply coefficient mutation to the offspring obtained after GOM has taken place, a single time.
    \item \emph{FOS-size times, after GOM}: 
    As explained in \Cref{sec:gom}, GOM performs many more changes (or better, attempts) than classic recombination operators, namely as many as the number of subsets contained in the FOS.
    With this strategy, we take this into account. 
    Like for the previous strategy, we apply coefficient mutation only after GOM has ended.
    Differently from before, we do not apply coefficient mutation once, but as many times as the number of attempts GOM makes (i.e., the number of subsets in the FOS).
    \item \emph{In between GOM steps}: A different strategy we consider is to apply coefficient mutation interleaved with the attempts GOM makes.
    Specifically, after a subset of the FOS has been used to attempt to improve the offspring, and the fitness evaluation has been carried out to accept or reject that attempt, then we apply coefficient mutation.
    This behavior is different from applying coefficient mutation only after all GOM steps have terminated (as per two the previous strategies) because changing a coefficient before a GOM step may impact whether that step will be successful.
    \item \emph{Within GOM steps}: The last strategy we consider applies coefficient mutation only to the coefficients that are considered \emph{within} a GOM step.
    Recall that, during a GOM step, the nodes identified by the subset in consideration are copied from a random donor solution into the offspring, replacing the existing ones.
    Now, if any of the nodes to be copied from the donor represents a coefficient, then coefficient mutation is applied.
    Thus, some of the coefficients being copied may be altered.
\end{enumerate}

For all the strategies, we follow the hill-climbing nature of GOM, i.e., we only keep coefficient mutation changes that do not cause the fitness to worsen (as per \Cref{alg:assess-changes}).
This means that after every coefficient mutation attempt, a fitness evaluation is needed.
The extra cost per offspring in terms of fitness evaluations for the different strategies is:
one for strategy (1), the size of the FOS (e.g., $2\ell -2$ for the LT) for strategies (2) and (3), and zero for strategy (3).
The latter follows from the fact that coefficient mutation is applied within the GOM step, before fitness evaluation takes place.
The idea of applying coefficient mutation as many times as the number of mixing attempts is inspired from previous work on model-based optimization of real-valued variables alongside discrete variables, which indicated that it is important to strike a proper balance between the number of discrete mixing events and the number of times the real values are sampled~\cite{sadowski2016learning,sadowski2018gambit}.

\section{Experimental setup}
\label{sec:experimental-setup}
We set up GP-GOMEA according to the hyper-parameter settings that are default or were found to work well in SRBench~\cite{la2021contemporary} by automatic hyper-parameter tuning, see \Cref{tab:hyperparamsetting}.
The only hyper-parameters we vary are those related to coefficient mutation, and the depth of the template tree. 
The latter is due to the fact that the number of evaluations in GP-GOMEA scales linearly with the number of nodes allowed for the trees (see \Cref{sec:gp-gomea}), thus using a smaller template tree means that the evolution can proceed for a longer time.

\begin{table}[]
    \caption{Hyper-parameter settings considered. 
    The top part of the table includes general hyper-parameters of GP-GOMEA, with all fixed settings except for template tree depth.
    The bottom part concerns hyper-parameters of coefficient mutation, which are all varied.
    }
    \centering
    \resizebox{\linewidth}{!}{%
    \begin{tabular}{lc}
    \toprule
        Hyper-parameter & Setting \\
    \hline
    \multicolumn{2}{c}{General} \\
    \hline
        Atomic functions & $\mathcal{F}=\{ +, -, \times, \div, \log, \sqrt{\cdot}, \sin, \cos \}$ \\
        Coefficient initialization & $\max_{i,j} | x^{(j)}_i | \times \mathcal{U}(-5,+5)$ \\
        Population initialization & Half-and-half \\
        Population size & \num{1000} \\
        Fitness function & Mean squared error \\
        FOS & LT \\
        Linear scaling~\cite{keijzer2003improving} & Active \\
        Template tree depth & 4 or 6 \\
        Termination criterion & \num{1000000} evaluations \\
    \hline
    \multicolumn{2}{c}{Coefficient mutation} \\
    \hline
        Probability & 0.5 or 0.9 \\
        Strength & \emph{ES-like} or \emph{Temp.} w/ $\tau=$\num{0.1} or \num{0.9}\\
        Decay (only \emph{Temp.}) & None, 0.1, or 0.9 \\
        Patience (only \emph{Temp.}) & Infinite or \num{5} generations \\
        Strategy & Never, after (1 or |FOS|), in between, or within \\
    \bottomrule
    \end{tabular}
    }
    \label{tab:hyperparamsetting}
\end{table}

We consider two experiments, which use different benchmark sets.
In the first experiment, we study the impact of coefficient mutation in all its hyper-parameter setting combinations, to understand what settings appear to be most relevant.
There, we consider a subset of the data sets collected in~\cite{oliveira2018analysing}.
In the second experiment, we apply the most promising setting to another set of problems, from SRBench~\cite{la2021contemporary}.

\section{Experiment 1: Hyper-parameter importance \& configuration}
\label{sec:first-exp}
We consider~\cite{oliveira2018analysing}, where the authors collected a list of data sets that were used in papers presented at the Genetic and Evolutionary Computation Conference (GECCO) from 2013 to 2017.
We particularly focus on the synthetic data sets, which are created by sampling relatively small ground-truth equations.
Of the 50 data sets listed in~\cite{oliveira2018analysing}, we consider those that were generated from equations that contain at least two coefficients, resulting in 23 data sets.

Since the considered ground-truth equations are relatively small, we use a template tree depth of $4$ for this experiment.
At the same time, we consider all setting combinations for the hyper-parameters concerning coefficient mutation (see~\Cref{tab:hyperparamsetting}).
For each combination and data set, we run GP-GOMEA ten times, to account for randomness.
The split between training and test set is pre-defined and data set-dependent~\cite{oliveira2018analysing}.
If a training set contains more than $256$ observations, we use a batch size of $256$ (randomized every generation) to speed up the experiments.

After having run GP-GOMEA, we attempt to infer what coefficient mutation hyper-parameters appear to be most important to determine GP-GOMEA's performance. 
We do this by (1) assembling a data set in which hyper-parameter settings represent features and the median training error achieved by ten runs of GP-GOMEA represents the label and (2) fitting a regression random forest~\cite{breiman2001random} to predict the median training error obtained by GP-GOMEA on each data set from the hyper-parameter settings used.
We consider the training error instead of the test error to focus on pure optimization performance; moreover, the test error provides a more noisy signal due to the generalization gap. 

\begin{figure}
    \centering
    \setlength\tabcolsep{0pt}
    \begin{tabular}{cc}
        \includegraphics[width=0.48\linewidth]{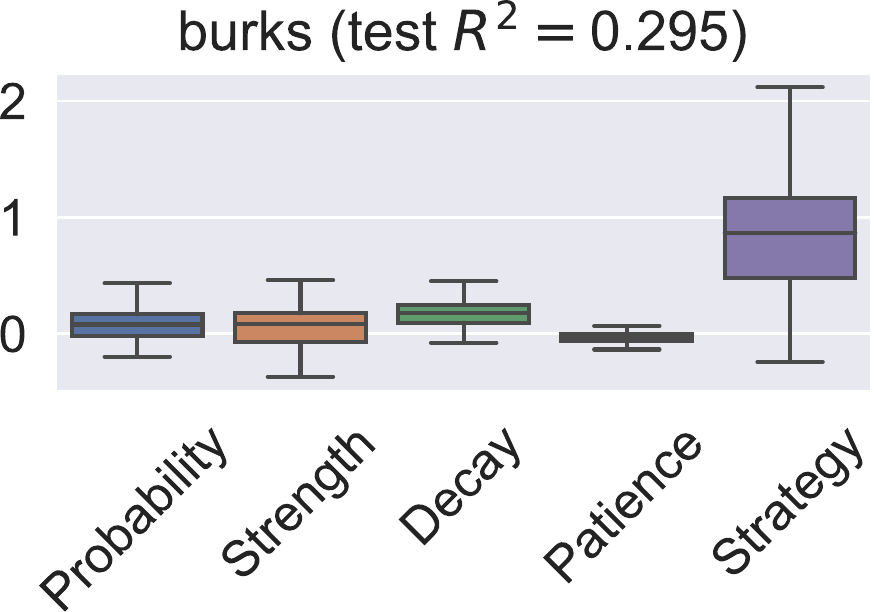}
        &
        \includegraphics[width=0.48\linewidth]{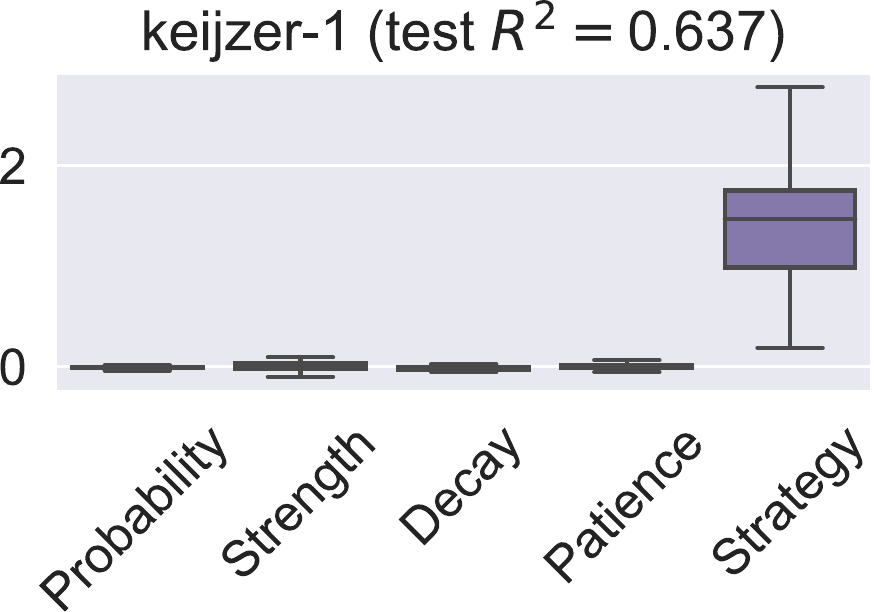}
        \\
        \includegraphics[width=0.48\linewidth]{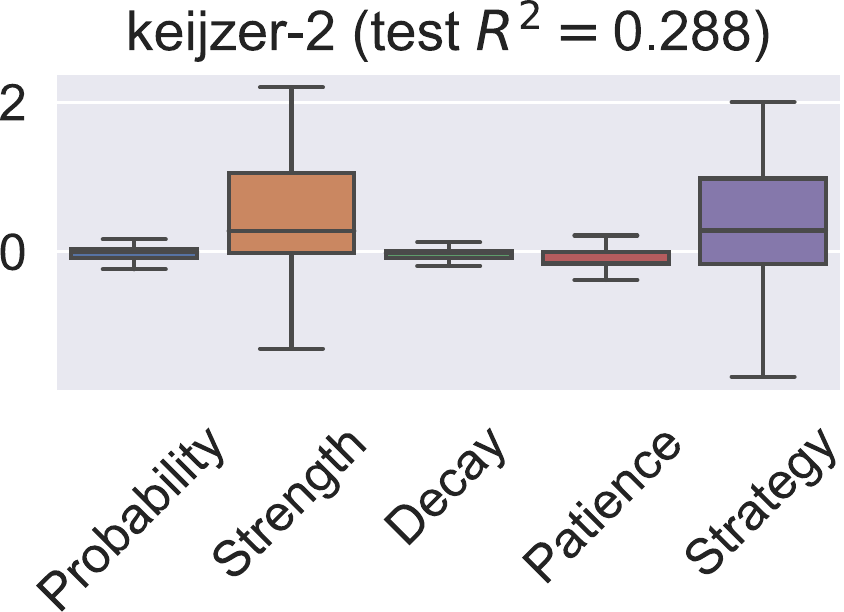}
        &
        \includegraphics[width=0.48\linewidth]{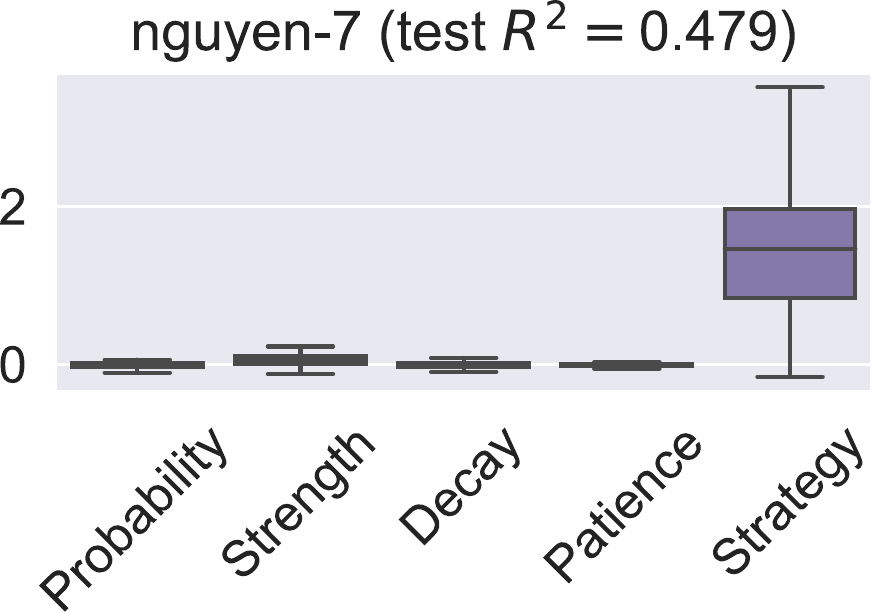}
        \\
        \multicolumn{2}{c}{\includegraphics[width=0.45\linewidth]{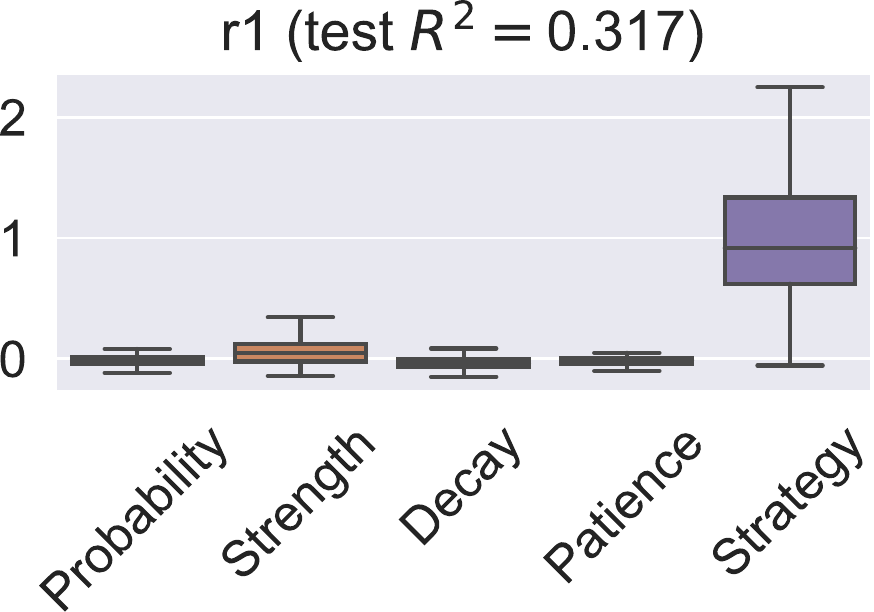}}
        \end{tabular}
    \caption{Hyper-parameter importance for the data sets upon which random forest learns a decent mapping (test $R^2 > 0.25$) between hyper-parameter setting and respective training performance of GP-GOMEA.
    Note that a negative values are due to random permutations used to estimate hyper-parameter importance leading to improved random forest test $R^2$ (less likely for good values of test $R^2 > 0.25$).}
    \label{fig:hyperparam_imp}
\end{figure}

To assess random forest's feature importance (and thus GP-GOMEA's hyper-parameter importance), for ten repetitions, we split the data set obtained at step (1) into 80\% training and 20\% testing; fit the forest on the training and measure the quality of fit on the test set (in terms of $R^2$-score); and, if such quality is decent (we impose that the test $R^2\text{-score} \geq 0.25$ as rule-of-thumb), use the permutation importance method~\cite{breiman2001random}, using ten more repetitions.
The result of this is displayed in \Cref{fig:hyperparam_imp}, for the five data sets where we found that random forest learned a meaningful mapping (test $R^2\text{-score} \geq 0.25$).
The general trend that can be observed is that, typically, the coefficient mutation strategy is the most important hyper-parameter.

\begin{figure*}
    \centering
    \begin{tabular}{ccc}
        \includegraphics[width=0.3\linewidth]{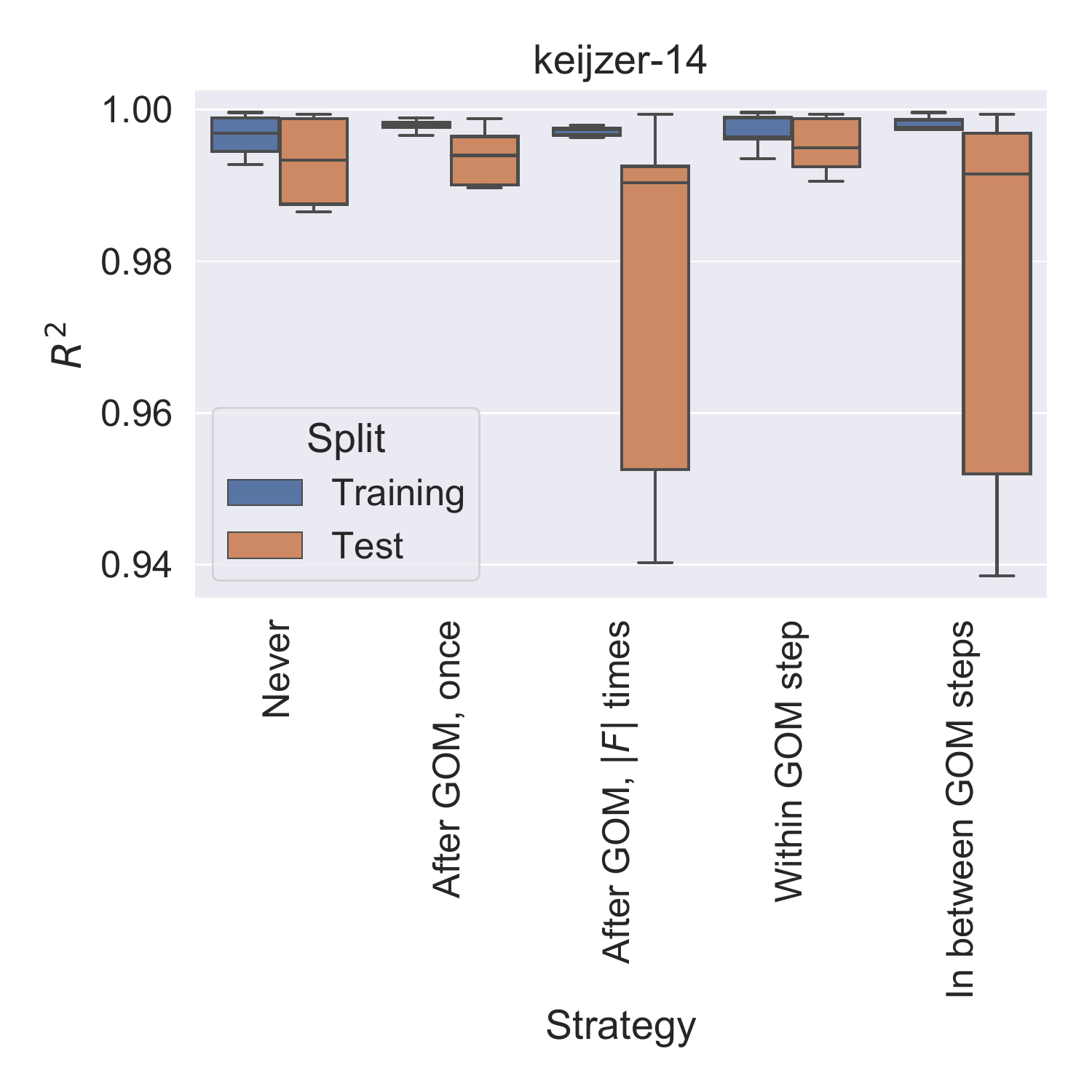}
        & 
        \includegraphics[width=0.3\linewidth]{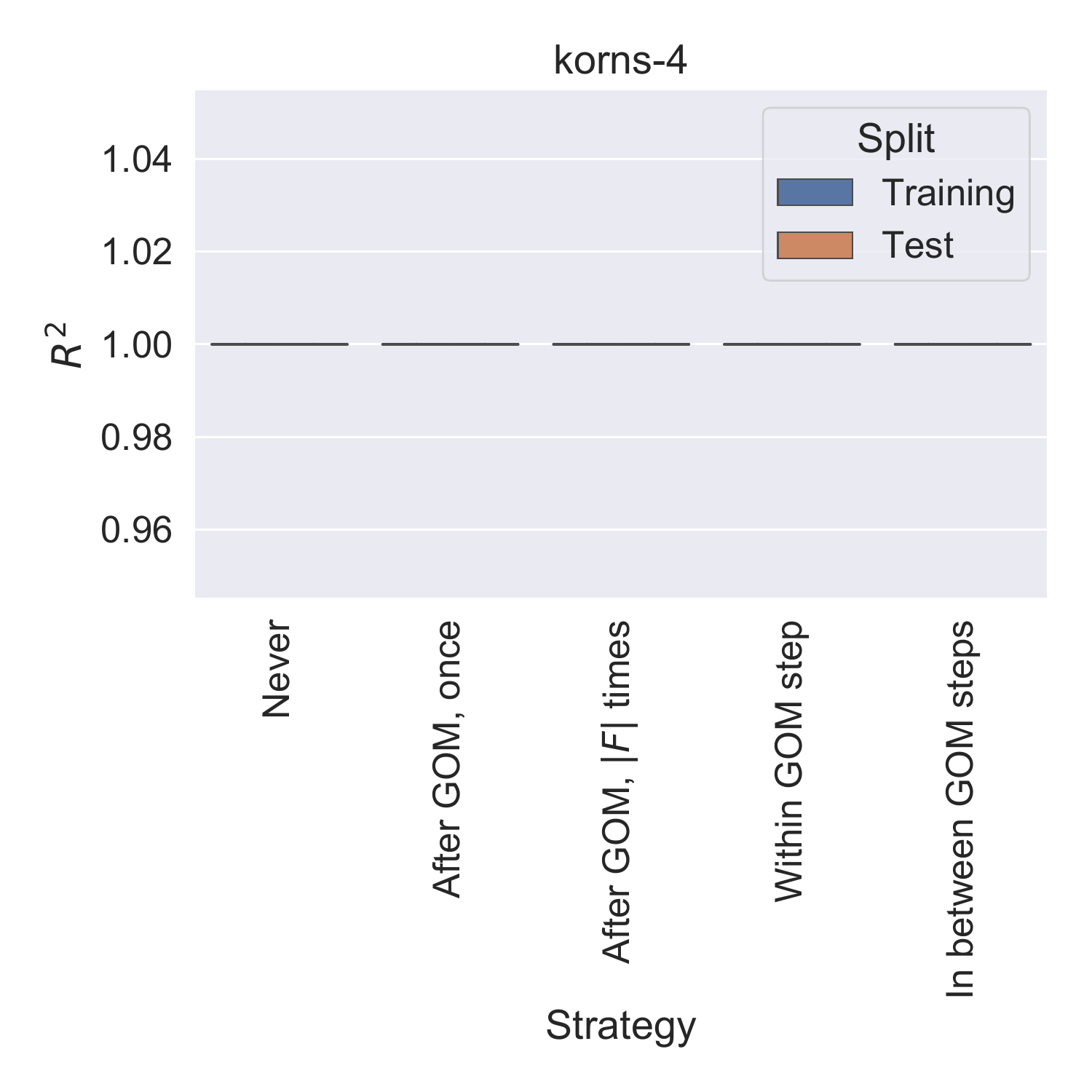}
        & 
        \includegraphics[width=0.3\linewidth]{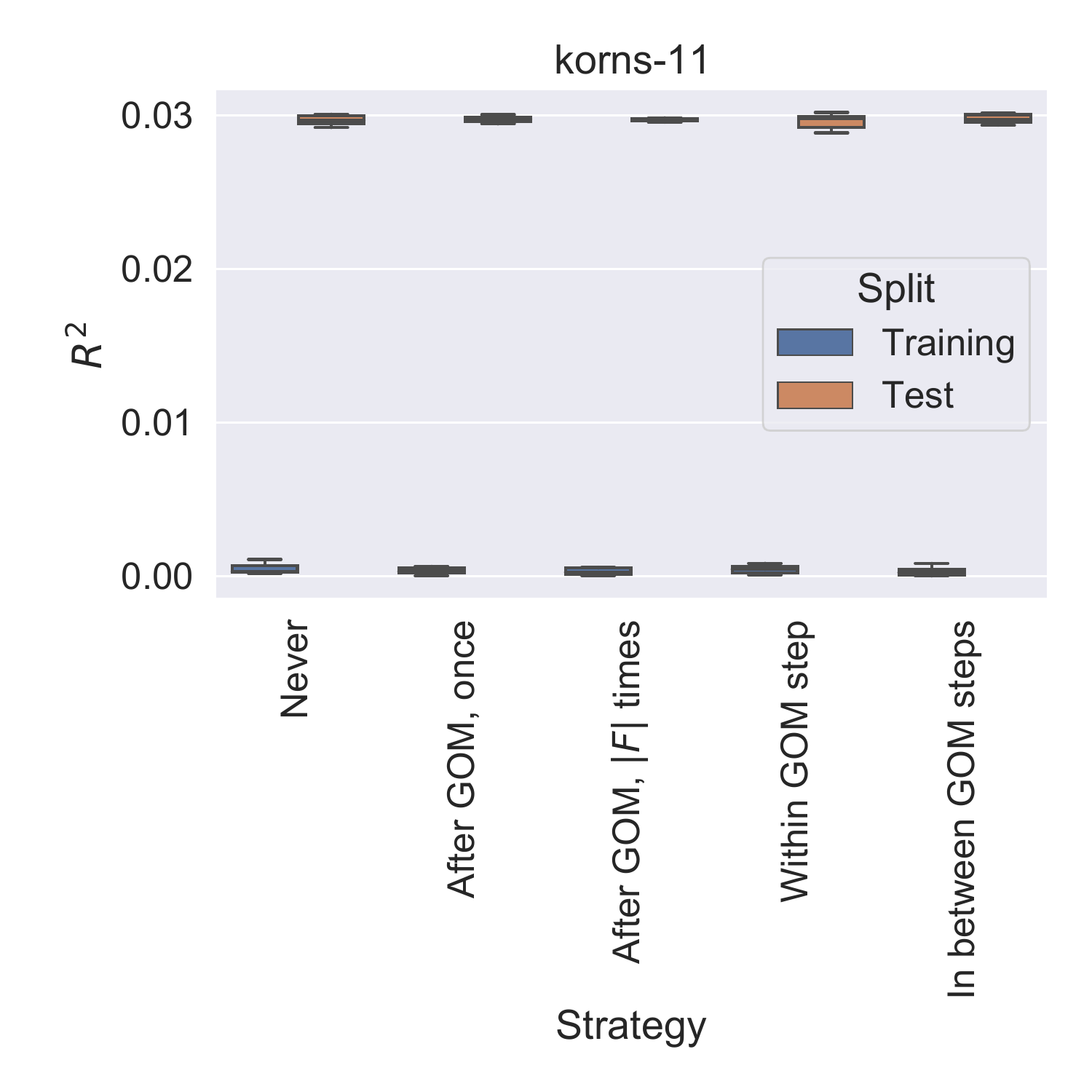}
    \end{tabular}
    \caption{
    GP-GOMEA's performance for three data sets for which it was not possible to learn how hyper-parameter settings influence (training) performance.
    The hyper-parameters are set to the best configuration found, expect for \emph{strategy}, which is varied.
    In all cases, the (training) performance differences are too small across settings.
    Left: the runs found relatively good solutions.
    Middle: the runs always found perfect solutions.
    Right: the runs never found good solutions.
    }
    \label{fig:performance_bad}
\end{figure*}

\begin{figure*}
    \centering
    \begin{tabular}{ccc}
        \includegraphics[width=0.3\linewidth]{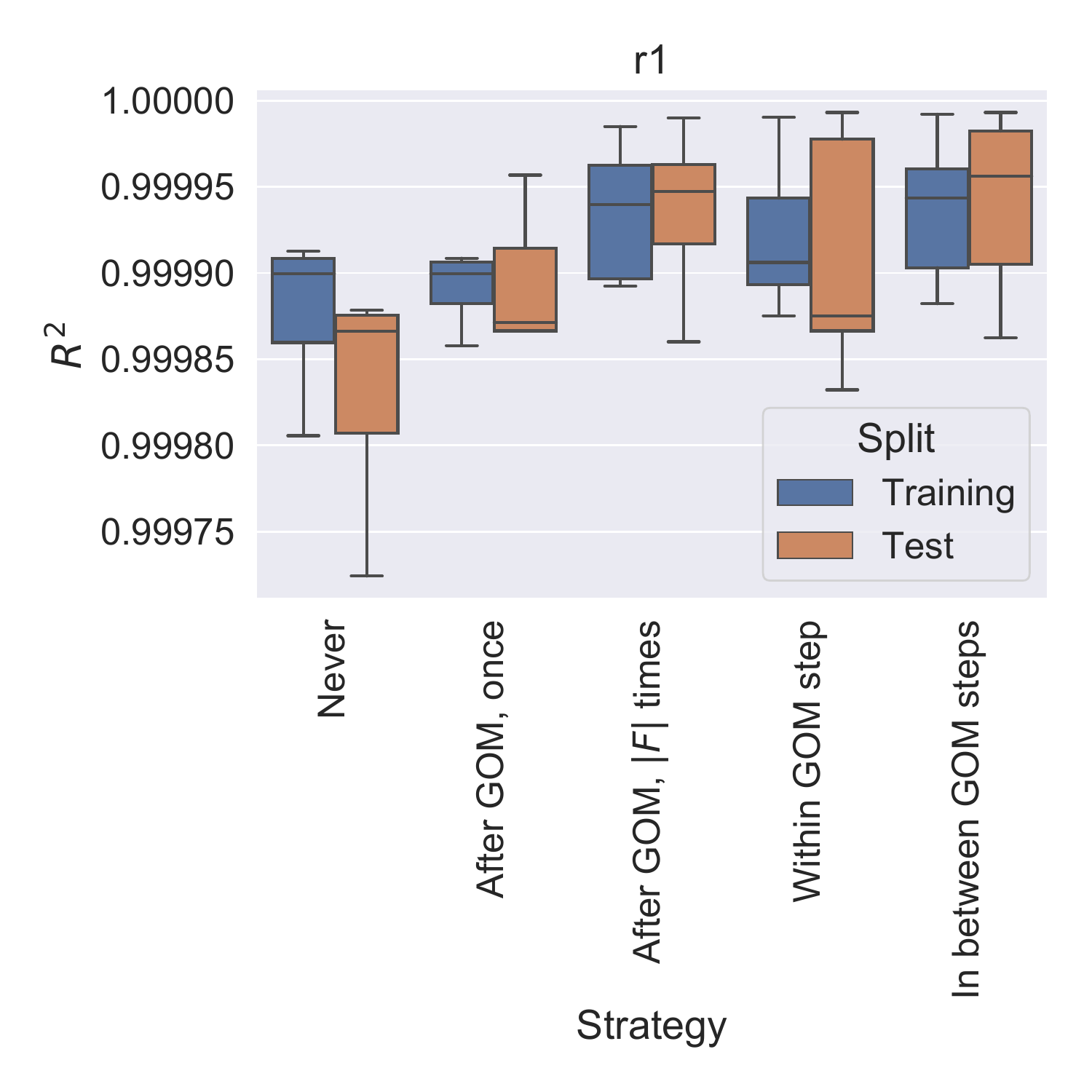}
        & 
        \includegraphics[width=0.3\linewidth]{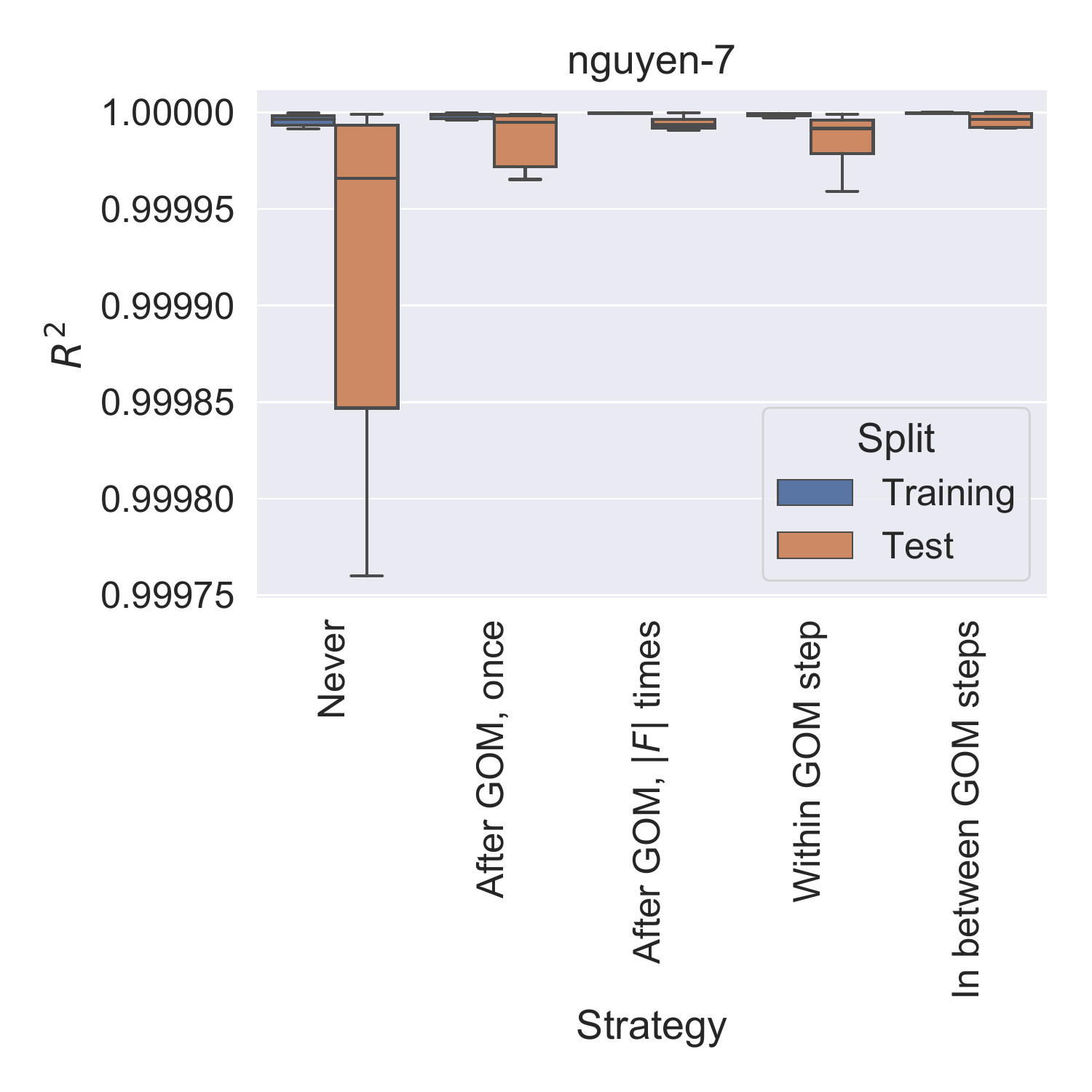}
        & 
        \includegraphics[width=0.3\linewidth]{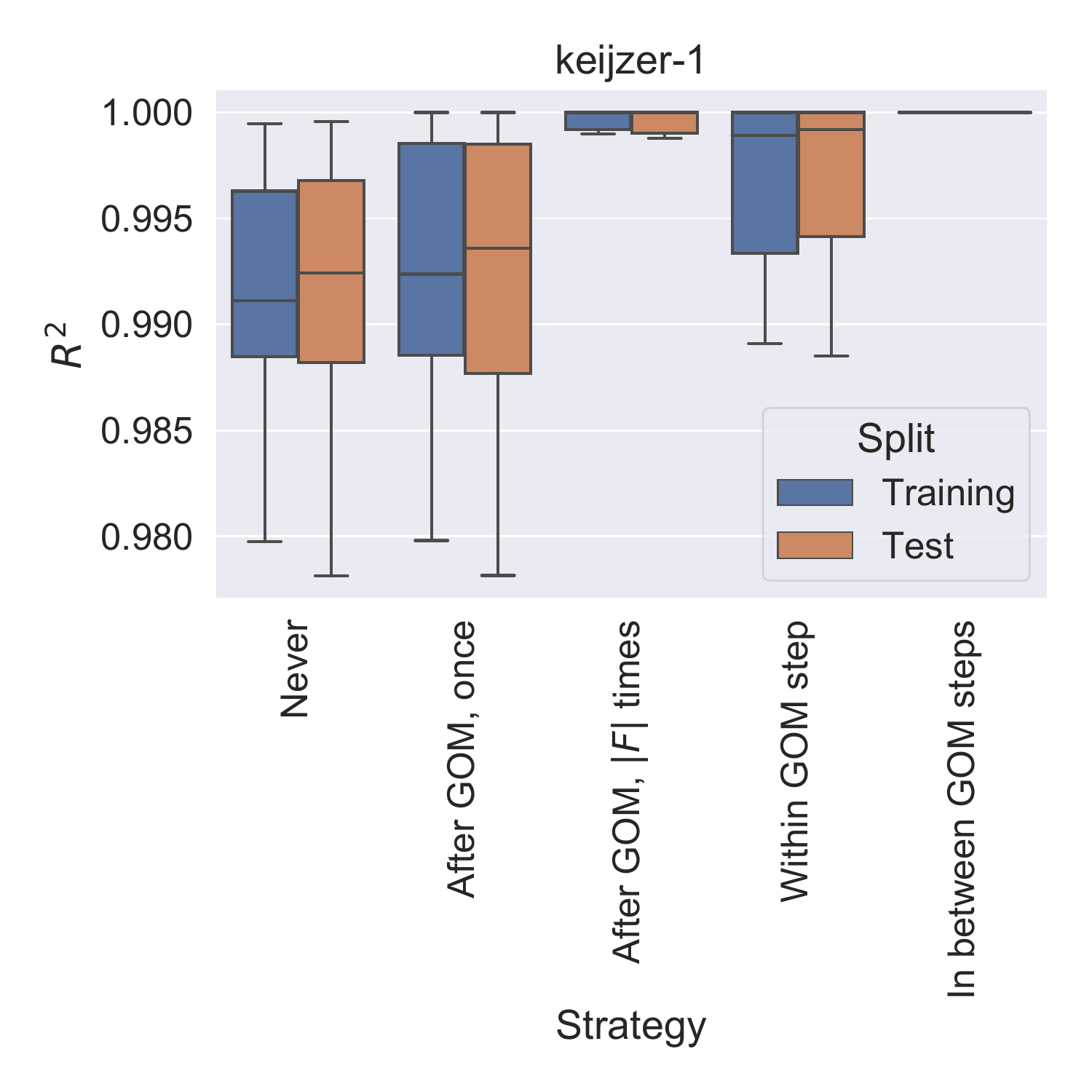}
    \end{tabular}
    \caption{
    GP-GOMEA's performance for three data sets for which it was possible to learn how hyper-parameter settings influence (training) performance.
    The hyper-parameters are set to the best configuration found, expect for \emph{strategy}, which is varied.
    }
    \label{fig:performance_ok}
\end{figure*}

Before proceeding, we provide examples of why random forest did not learn a meaningful mapping between GP-GOMEA's coefficient mutation hyper-parameter setting and its training performance for most data sets.
Since the coefficient mutation strategy is the most important hyper-parameter from \Cref{fig:hyperparam_imp}, we keep this hyper-parameter variable, while we fix the others, to a good setting (we explain how this is obtained in the next paragraph), namely probability of $1$, strength temperature-based with $\tau=0.1$, decay of $0.1$ with patience of $5$ generations. 
Now,~\Cref{fig:performance_bad} shows how performance changes according to the strategy when the other hyper-parameters are fixed as just mentioned, for three of the data sets for which the random forest's test $R^2 < 0.25$.
As it can be seen, GP-GOMEA obtains limited differences in performance when varying the strategy of coefficient mutation for different reasons.
Conversely, \Cref{fig:performance_ok} shows that the strategy has an impact on three of the data sets for which random forest obtained $R^2 \geq 0.25$.

\begin{figure*}
    \centering
    \setlength\tabcolsep{0pt}
    \begin{tabular}{cccccc}
        \includegraphics[width=0.2571\linewidth]{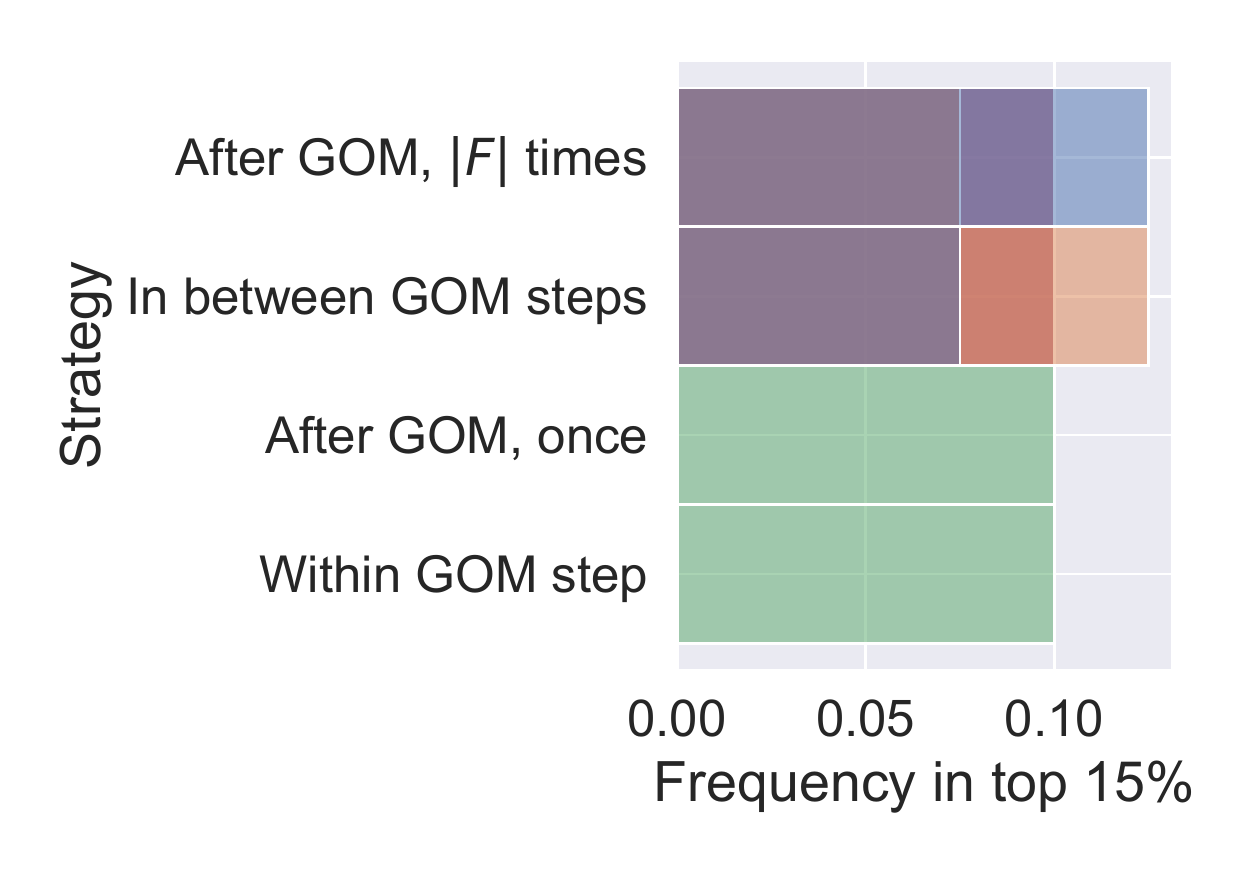} &
        \includegraphics[width=0.18\linewidth]{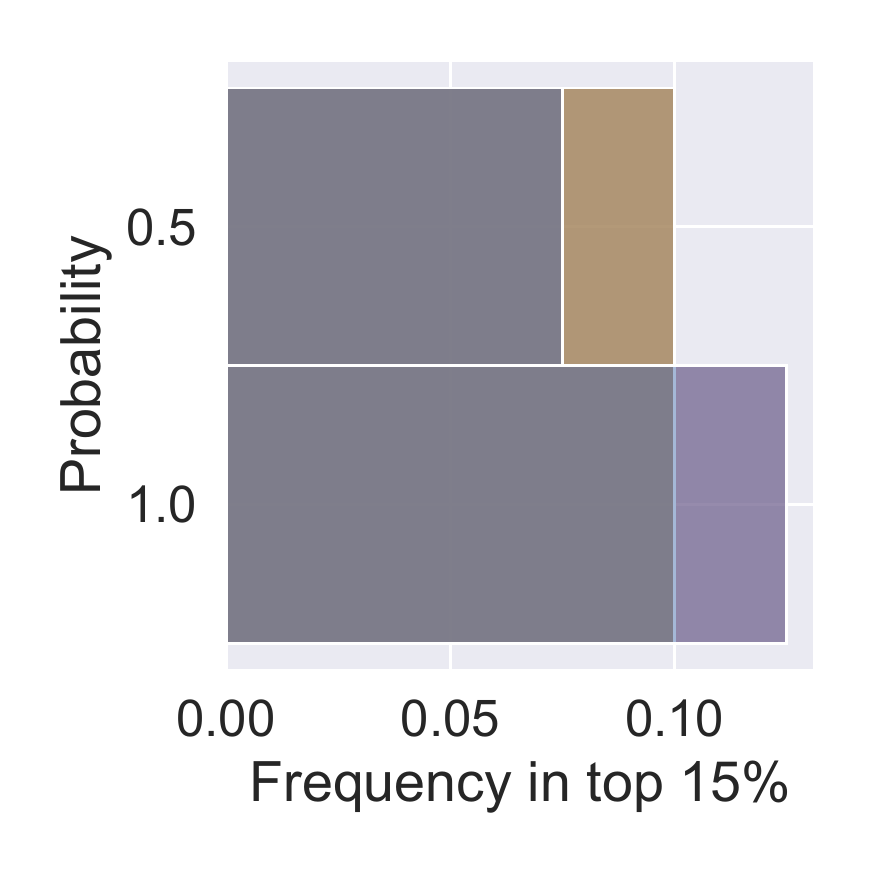} &
        \includegraphics[width=0.18\linewidth]{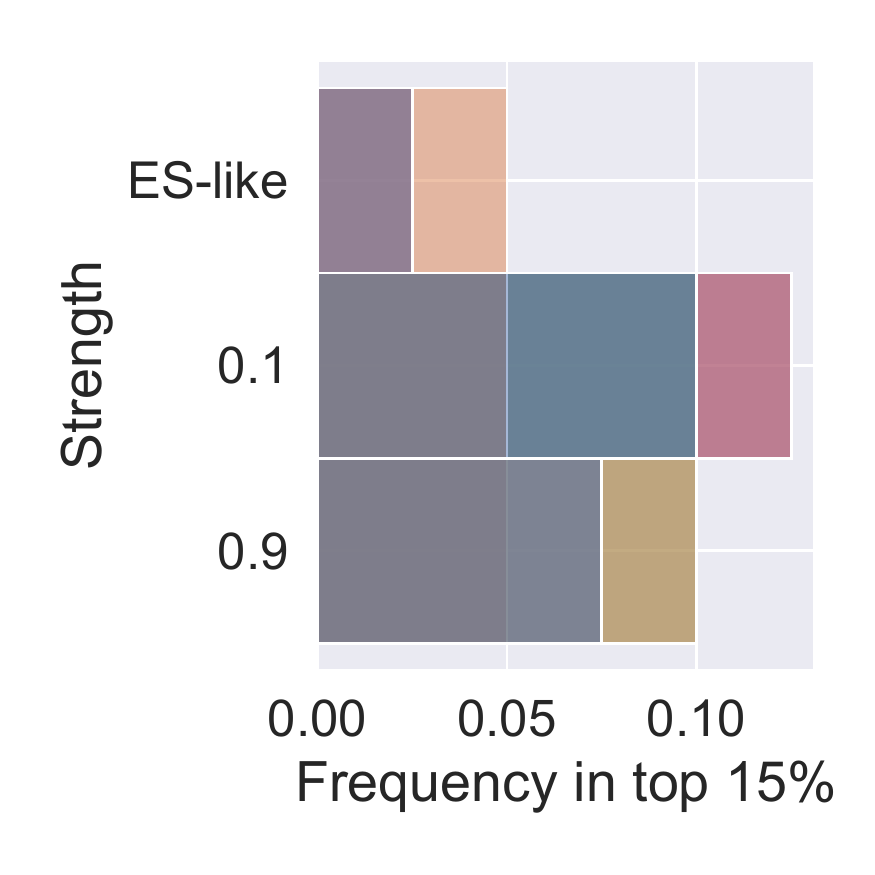} &
        \includegraphics[width=0.18\linewidth]{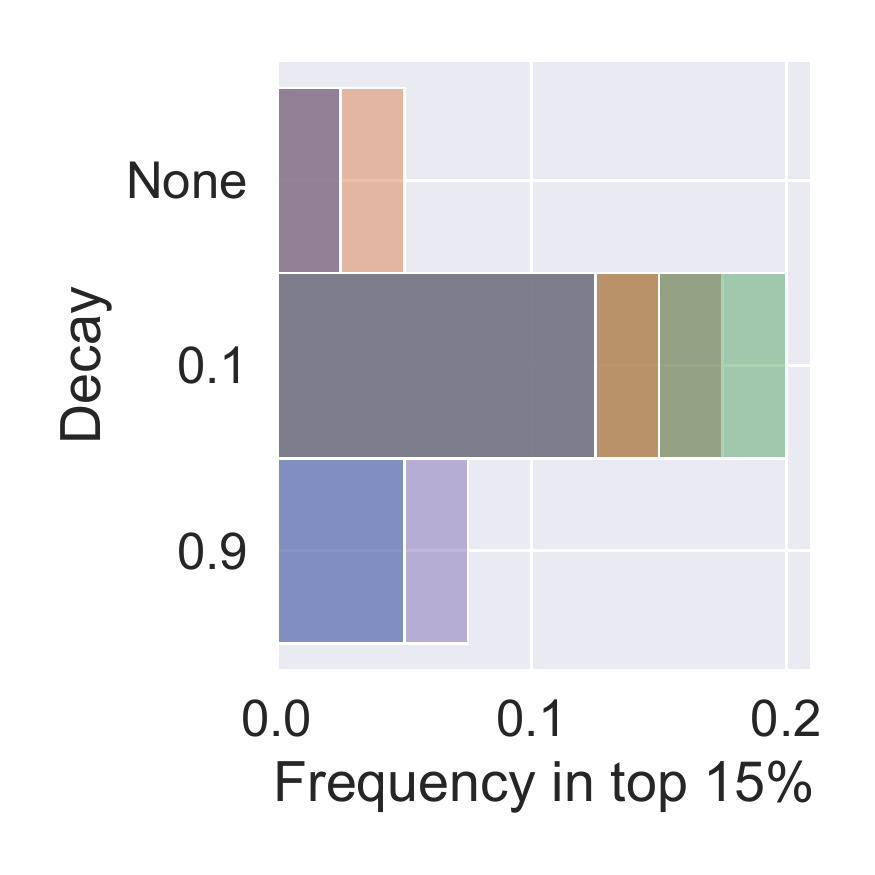} &
        \includegraphics[width=0.18\linewidth]{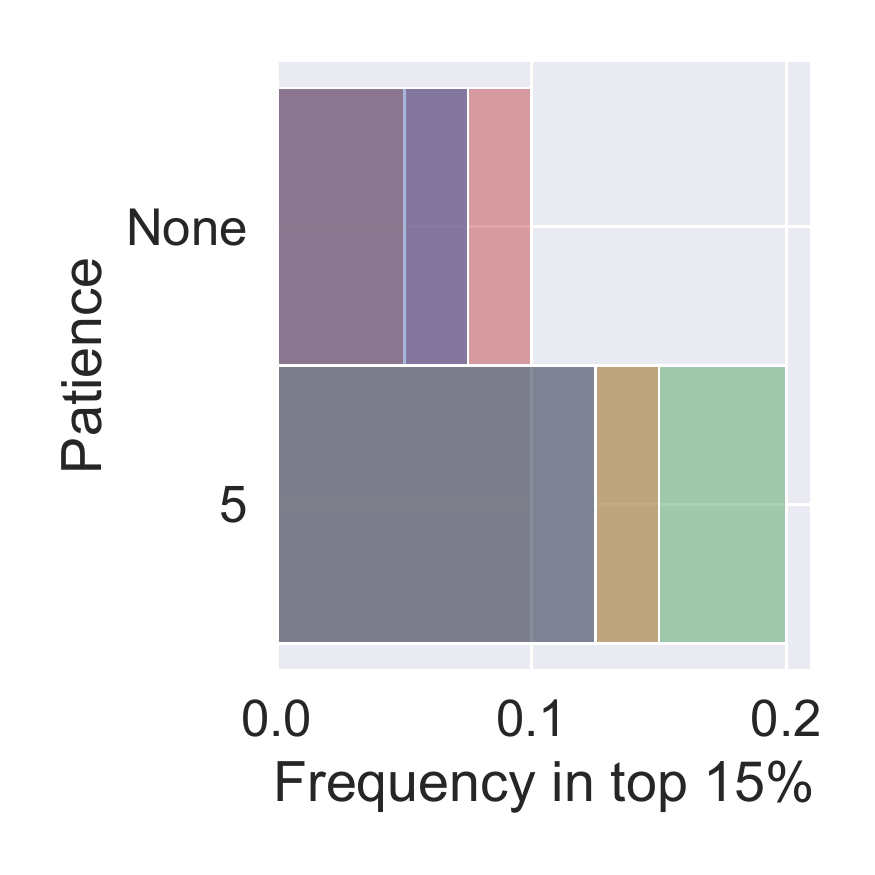} \\
        %\multicolumn{5}{c}{\includegraphics[height=2cm]{pics/importance/top_performing_legend.pdf}}
    \end{tabular}
    \caption{Frequency with which a coefficient mutation hyper-parameter setting appears in the top 15\% of training performance, for the data sets {\color{NavyBlue}burks}, {\color{BurntOrange}keijzer-1}, {\color{OliveGreen}keijzer-2}, {\color{BrickRed}nguyen-7}, and {\color{Purple}r1}. }
    \label{fig:top-performing}
\end{figure*}

\begin{figure}
    \centering
    \includegraphics[width=\linewidth]{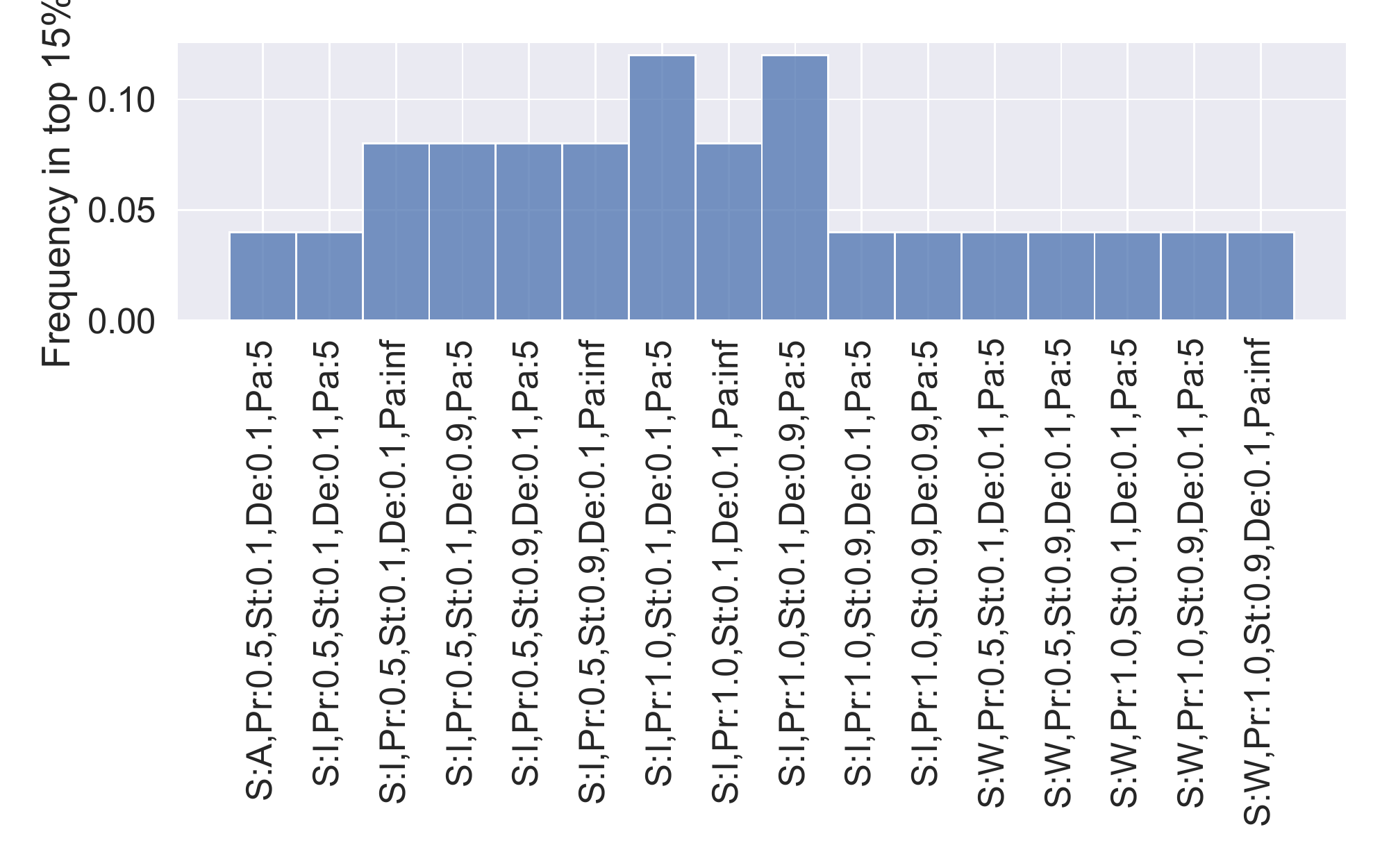}
    \caption{Frequency with which a hyper-parameter configuration appears among the top 15\% of training performance obtained by GP-GOMEA. 
    The abbreviations \emph{S:A}, \emph{S:I}, and \emph{S:W} stand for \emph{strategy} set to \emph{after GOM |FOS| times}, \emph{in between GOM steps}, and \emph{within GOM steps}, respectively.
    The other abbreviations stand for, in order, \emph{probability}, \emph{strength}, \emph{decay}, and \emph{patience}.
    }
    \label{fig:top-performing-joint}
\end{figure}

Finally, we consider the five data sets of \Cref{fig:hyperparam_imp} and we look at the hyper-parameter settings that lead to top $15\%$ training performance.
The result is shown separately for hyper-parameter and data set in \Cref{fig:top-performing}, and jointly across both in \Cref{fig:top-performing-joint}.
Two configurations appear to be most promising, which have in common the \emph{in between GOM steps} strategy, probability of $1$, strength temperature-based with $\tau=0.1$, and patience of $5$ generations, while decay can be set to $0.1$ or $0.9$. 
We pick the decay of $0.1$ and proceed with the next experiment.

\section{Experiment 2: Application to SRBench}
\label{sec:second-exp}
We use the promising hyper-parameter configuration from the previous experiment and proceed with benchmarking GP-GOMEA with coefficient mutation on the so-called \emph{ground-truth} data sets of SRBench.
These data sets were generated from a known, ground-truth equation, with varying level of noise added to the target variable (see~\cite{la2021contemporary} for details).
Here, we evaluate two options for the depth of the template tree (hereon simply referred to as \emph{depth}, for brevity), i.e., $4$ and $6$, and two options for coefficient mutation, i.e., active (using the promising configuration) or inactive.
Note that the current results for GP-GOMEA reported in~\cite{la2021contemporary} and on the repository of SRBench use a depth of $6$ and no coefficient mutation.

\begin{figure}
    \centering
    \includegraphics[width=0.9\linewidth]{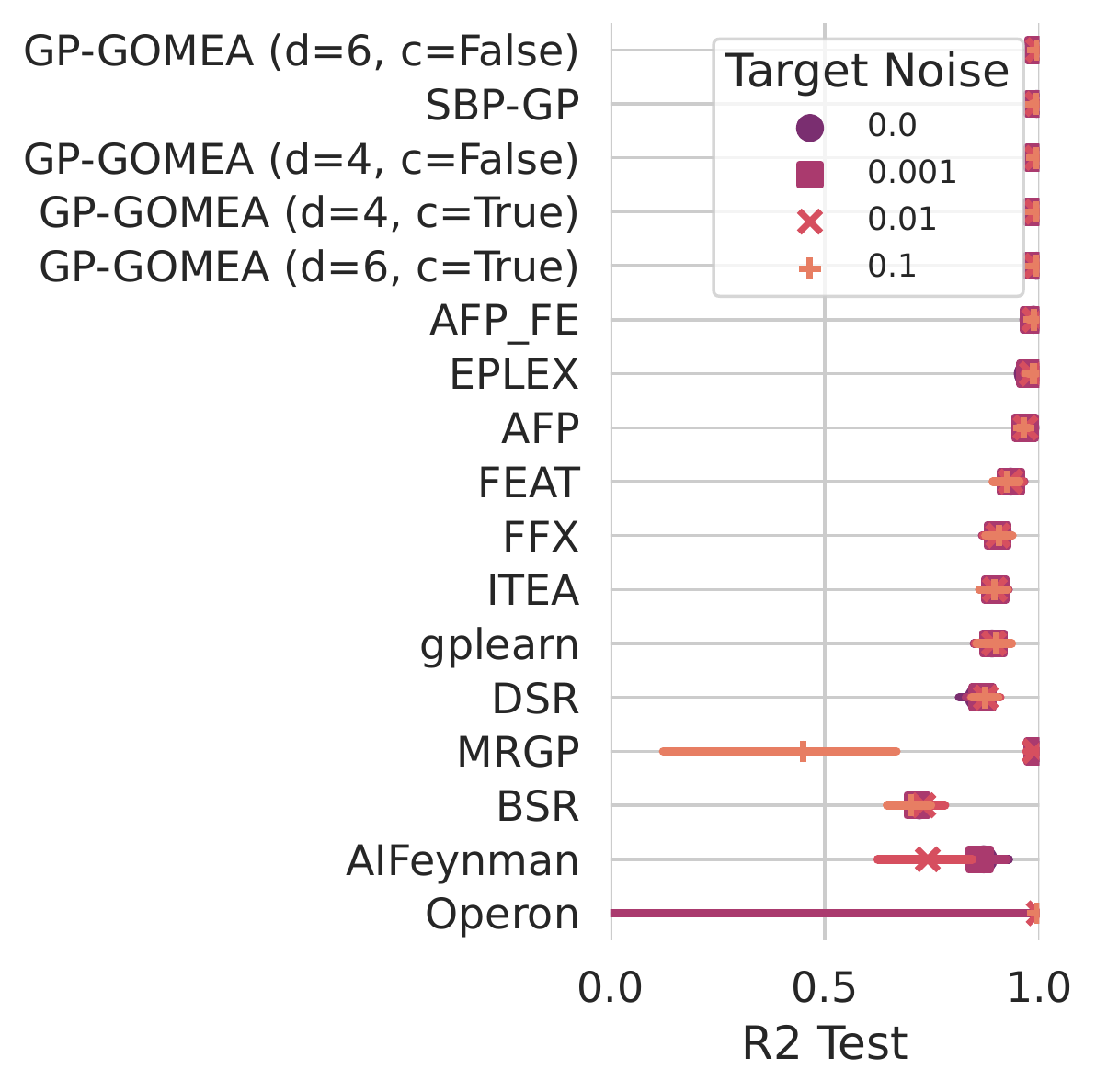}
    \caption{Test $R^2$ for the variants of GP-GOMEA and other algorithms currently benchmarked in SRBench.
    The large error bar of Operon is due to an outlier.}
    \label{fig:srbench_acc}
\end{figure}

\begin{figure}
    \centering
    \includegraphics[width=0.9\linewidth]{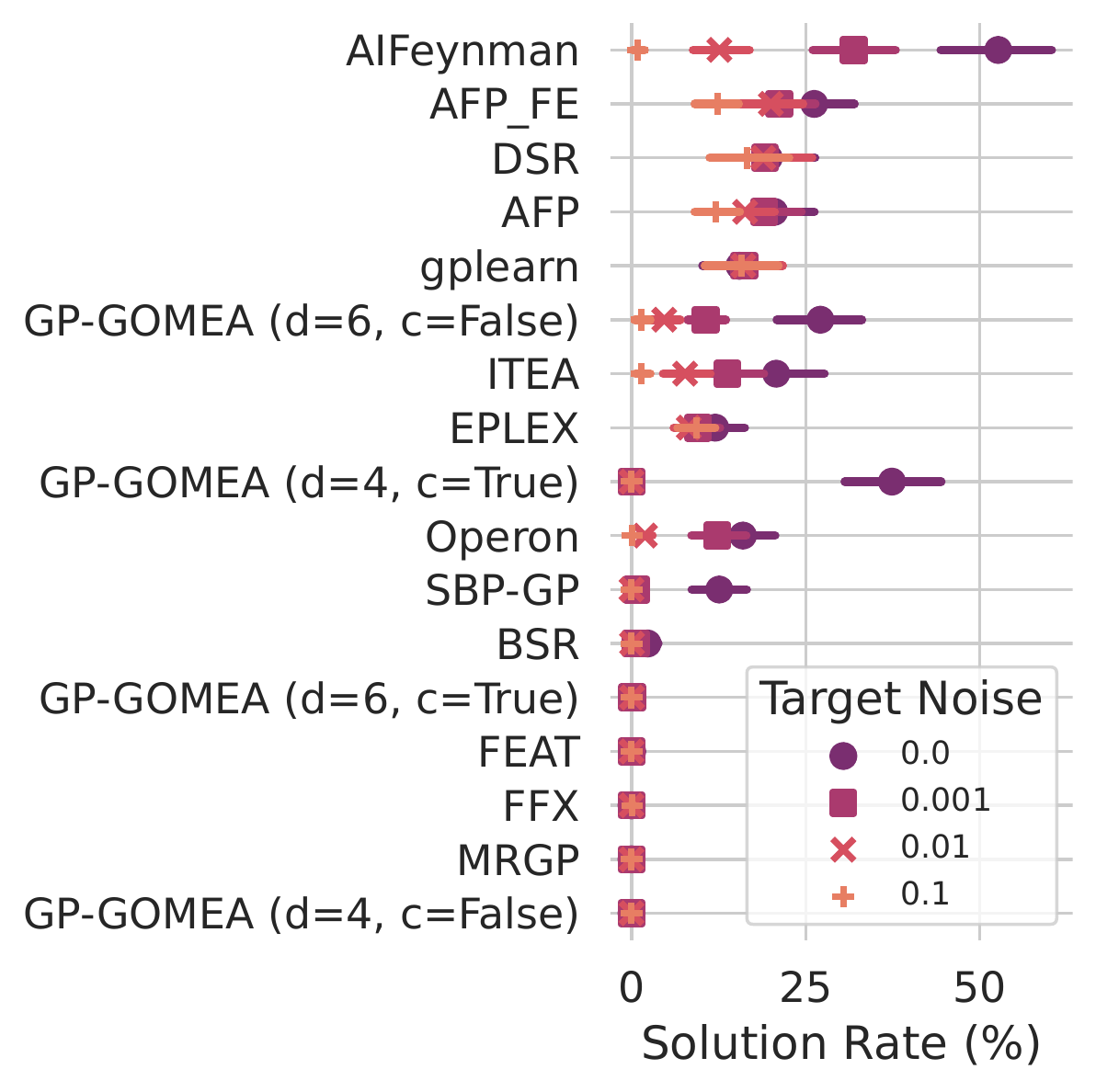}
    \caption{Rate with which an algorithm discovers the ground-truth equation from which the data set was generated, for the variants of GP-GOMEA and other algorithms currently benchmarked in SRBench.}
    \label{fig:srbench_symsol}
\end{figure}

We begin by considering the test $R^2$ obtained by the evolved models, as shown in \Cref{fig:srbench_acc} (see~\cite{la2021contemporary} for full names and descriptions of the competing algorithms).
All four configurations of GP-GOMEA perform very competitively, with rather small differences in terms of $R^2$ across different noise levels.
Coefficient mutation seems not to make a large difference here.
To get a more complete view, in \Cref{fig:srbench_symsol} we report the \emph{solution rate}, i.e., the frequency (out of ten repetitions) with which the algorithms re-discover the ground-truth equations from which the data set was generated.
There, two configurations of GP-GOMEA achieve better results, namely the one using a depth of \num{6} and inactive coefficient mutation, and the one using a depth of \num{4} and active coefficient mutation.
Regarding the other two configurations, using a depth of 6 and active coefficient mutation perform worse because it requires the largest number of evaluations (SRBench uses a budget of \num{1000000} evaluations), while using a depth of 4 and inactive coefficient mutation makes it hard for GP-GOMEA to refine the models, which are necessarily small due to the constrained template size.

Interestingly, despite having a smaller template for representing solutions, GP-GOMEA with depth of 4 and active coefficient mutation can be competitive with GP-GOMEA with depth of 6 and inactive coefficient mutation, at least when no noise is added to the target variable.
Actually, without noise, coefficient mutation allows to substantially improve the discovery of the ground-truth equation, by approximately $+10\%$ (compared to GP-GOMEA with depth 6 and inactive coefficient mutation).
However, when noise is added to the target, coefficient mutation makes GP-GOMEA dramatically less reliable in discovering the ground-truth equation.

\section{Discussion}
\label{sec:discussion}

As variation in GP-GOMEA works differently than in GP, we investigated different strategies that determine at what stage and for how many attempts coefficient mutation should be applied in GP-GOMEA.
In our first experiment (\Cref{sec:first-exp}), we found that the choice of this strategy is typically the most important factor at play; at least for the data sets in which coefficient mutation plays a substantial role.
The best strategy we found is to apply coefficient mutation in between every step of GOM.
However, good results were also observed when applying coefficient mutation after all GOM steps had taken place, as long as the number of attempts matched those of GOM (i.e., the size of the FOS).
In particular, applying coefficient mutation only a single time after an offspring is generated is not sufficient in GP-GOMEA.

Other hyper-parameter settings concerning coefficient mutation are generally less important.
The temperature-based way of determining the strength of coefficient mutation, at least for the experimental setup we used (e.g., with the evaluation budget of SRBench), performed slightly better than the ES-like approach. 
However, the ES-like approach has the appeal that it requires less hyper-parameters to be set, and is likely to reach similar performance if more budget is given thanks to its self-adaptation.
In particular, for this approach, it suffices that $\epsilon$ is a small number (we used $10^{-16}$), but a better choice of $\gamma$ may be important (we used $0.1$).
We remark that the orders of magnitude for the sampling variance used in the ES-like approach and in the temperature-based approach at initialization can be dissimilar but not wildly so.
With our choice of $\gamma$, we obtained that, at initialization, the sampling variance is approximately $1$ for all constant nodes (see \Cref{eq:sigma,eq:es-update-c}).
For the temperature-based approach the sampling variance depends on $\tau$ and the current value of the coefficient $c$ (which is initialized according to the formula in \Cref{tab:hyperparamsetting}).
For a constant initialized at, e.g., $-5$ or $+10$, then using $\tau=0.1$ (found to perform best in \Cref{sec:first-exp}) leads to a sampling variance of $0.25$ or $1$, respectively.

A downside of coefficient mutation (compared to, e.g., gradient descent) is that changes may often be detrimental.
Therefore, one needs to evaluate whether the quality of a solution does improve after coefficient mutation, and roll back detrimental changes (at least in a stochastic manner). 
Under the assumption that node-wise recombination plays a key role in GP(-GOMEA), we did not investigate the effect of having more evaluations being spent for coefficient mutation instead of recombination (GOM). 
However, it may be important to study such scenarios, and of course to include other approaches for coefficient optimization. 
(Stochastic) gradient descent is a prime candidate for the future studies.

To understand the dynamics of coefficient mutation with respect to noisy measurements and discovery of ground-truth equations, we must consider
\Cref{fig:srbench_acc} and \Cref{fig:srbench_symsol} at the same time.
This comparison reveals that, when noise is present, GP-GOMEA with (depth 4 and) active coefficient mutation discovers decently accurate models that, however, do not match the ground-truth equations.
So, on the one hand, coefficient mutation allows for good models to be found under a relatively restrained representation (template tree with depth 4 compared to 6). 
On the other hand, coefficient mutation makes GP-GOMEA more susceptible to overfitting to noise.
This begs the question as to whether the use of coefficient mutation (or coefficient optimization in general) should be accompanied with the use of some form of regularization, to reduce the chance of overfitting to (noise in) the training set.

\section{Conclusion}
\label{sec:conclusion}
Tuning coefficients can be important to tackle symbolic regression effectively.
In this paper, we investigated how some simple, gradient-free forms of coefficient optimization can be integrated in GP-GOMEA, a state-of-the-art algorithms for symbolic regression.
We have carried out two sets of experiments, on two different benchmark sets.
In the first experiment, we have found that coefficient mutation does not always make a difference; however, when it does, then its most important factor is the strategy used to apply it.
In GP-GOMEA, the amount of coefficient mutation attempts needs to be commensurate to the amount of recombination attempts.
In the second experiment, we applied different variants of GP-GOMEA with and without coefficient mutation to the part of the SRBench benchmark that concerns discovering ground-truth equations from data.
We have found that coefficient mutation enhances GP-GOMEA's discovery success rate only if the data does not include noise.
With noisy data, coefficient mutation leads to finding different equations than the ground-truth ones, which however are similarly accurate.

%\begin{acks}
%To Robert, for the bagels and explaining CMYK and color spaces.
%\end{acks}

\bibliographystyle{ACM-Reference-Format}
\bibliography{paper}

\end{document}